\documentclass[sigconf]{aamas} 
\subtitle{Blue Sky Ideas Track}

\usepackage[inline]{enumitem}
\usepackage{mystyle}
\usepackage{soul}
\usepackage{breqn}
\usepackage{orcidlink}

\usepackage{balance} 

\newcommand{\problem}{Compositional Repertoire Learning}
\newcommand{\abbrev}{CRL}

\makeatletter
\gdef\@copyrightpermission{
  \begin{minipage}{0.2\columnwidth}
   \href{https://creativecommons.org/licenses/by/4.0/}{\includegraphics[width=0.90\textwidth]{by}}
  \end{minipage}\hfill
  \begin{minipage}{0.8\columnwidth}
   \href{https://creativecommons.org/licenses/by/4.0/}{This work is licensed under a Creative Commons Attribution International 4.0 License.}
  \end{minipage}
  \vspace{5pt}
}
\makeatother

\setcopyright{ifaamas}
\acmConference[AAMAS '26]{Proc.\@ of the 25th International Conference
on Autonomous Agents and Multiagent Systems (AAMAS 2026)}{May 25 -- 29, 2026}
{Paphos, Cyprus}{C.~Amato, L.~Dennis, V.~Mascardi, J.~Thangarajah (eds.)}
\copyrightyear{2026}
\acmYear{2026}
\acmDOI{}
\acmPrice{}
\acmISBN{}

\acmSubmissionID{13}

\title{%
    \texorpdfstring{%
        Beyond Mimicry: Toward Lifelong Adaptability \\ in Imitation Learning
    }{%
        Beyond Mimicry: Toward Lifelong Adaptability in Imitation Learning
    }
}

\author{Nathan Gavenski \orcidlink{0000-0003-0578-3086}}
\affiliation{
    \institution{King's College London}
    \city{London}
    \country{United Kingdom}}
\email{nathan.schneider_gavenski@kcl.ac.uk}
\orcid{0000-0003-0578-3086}

\author{Felipe Meneguzzi \orcidlink{0000-0003-3549-6168}}
\affiliation{
  \institution{University of Aberdeen \& PUCRS}
  \city{Aberdeen}
  \country{United Kingdom}}
\email{felipe.meneguzzi@abdn.ac.uk}
\orcid{0000-0003-3549-6168}

\author{Odinaldo Rodrigues \orcidlink{0000-0001-7823-1034}}
\affiliation{
  \institution{King's College London}
  \city{London}
  \country{United Kingdom}}
\email{odinaldo.rodrigues@kcl.ac.uk}
\orcid{0000-0001-7823-1034}

\begin{abstract}
Imitation learning stands at a crossroads: despite decades of progress, current imitation learning agents remain sophisticated memorisation machines, excelling at replay but failing when contexts shift or goals evolve.
This paper argues that this failure is not technical but foundational: imitation learning has been optimised for the wrong objective.
We propose a research agenda that redefines success from perfect replay to compositional adaptability.
Such adaptability hinges on learning behavioural primitives once and recombining them through novel contexts without retraining.
We establish metrics for compositional generalisation, propose hybrid architectures, and outline interdisciplinary research directions drawing on cognitive science and cultural evolution.
Agents that embed adaptability at the core of imitation learning thus have an essential capability for operating in an open-ended world.
\end{abstract}

\keywords{Imitation Learning, Generalisation, Compositional Learning}

\newcommand{\BibTeX}{\rm B\kern-.05em{\sc i\kern-.025em b}\kern-.08em\TeX}

\begin{document}


\pagestyle{fancy}
\fancyhead{}


\maketitle 


\section{Introduction}\label{sec:introduction}

When a child observes an adult stacking blocks, they do not memorise every motion in a trajectory.
Instead, they extract reusable behavioural primitives (grasping, lifting, placing) and the compositional rules that govern their combination.
Later, when faced with blocks of different shapes, sizes, or materials, a child can build entirely new structures never seen in demonstrations.
This compositional understanding, fundamental to human imitation learning (IL), remains conspicuously absent from modern IL systems.

The disconnect between human and artificial imitation reveals a foundational flaw in the field's approach to learning from demonstrations.
Current IL methods excel at reproducing demonstrated trajectories with impressive fidelity~\cite{gavenski2022how,ElhadadMeneguzziMirsky2026}.
Behavioural cloning (BC) achieves high accuracy on training distributions~\cite{pomerleau1988alvinn}, inverse reinforcement learning infers underlying objectives~\cite{gavenski2024survey}, and adversarial approaches can handle imperfect demonstrations~\cite{torabi2019recent}.
Yet, these successes mask a critical limitation: when contexts shift beyond the training distribution performance degrades significantly~\cite{gavenski2025labyrinth}.
An IL agent that masters a navigation layout cannot easily adapt to a configuration change.
These failures are not edge cases, but the norm whenever demonstration and evaluation conditions vary.
The brittleness stems from optimising for the wrong objective: sample efficiency rather than adaptability.
However, no amount of memorisation enables the behaviour required for more nuanced tasks.
The real challenge, we argue, is not simply to learn from fewer demonstrations, but to actually understand the compositional structure of tasks embedded in them.

This paper introduces \problem{} (\abbrev{}), a research agenda that redefines success in IL from trajectory reproduction to compositional generalisation.
Rather than treating demonstrations as sequences of actions to memorise, \abbrev{} requires agents to extract behavioural primitives and compositional rules that can be recombined to solve novel contexts.
This shift in orientation of the field is not merely technical but philosophical, fundamentally changing what it means for an agent to imitate successfully.
The core insight driving \abbrev{} is that demonstrations contain two types of information: surface trajectories and deep compositional structure.
Current methods capture only superficial knowledge, learning associations between states and actions.
\abbrev{} requires agents to capture the structure of the solution, understanding that they must grasp {\em primitives}, i.e., steps that achieve sub-goals, so they can be recombined according to task-dependent rules.
This structural understanding enables agents to generalise far beyond their training distribution, adapting to novel contexts through creative recombination rather than nearest-neighbour retrieval.

\section{Background} \label{sec:background}

Most IL works model environments as \textit{Markov Decision Processes} (MDPs), which are tuples of the form $\mdp = \langle\StateMDP, \Action, \Transition, \reward, \discount\rangle$, where $\StateMDP$ is the state space, $\Action$ is the action space, $\Transition: \StateMDP \times \Action \rightarrow \StateMDP$ is a transition function, $\reward: \StateMDP \times \Action \rightarrow \mathbb{R}$ is an immediate reward function, and $\discount$ is the discount factor~\cite{sutton2018reinforcement}.

\vskip 3pt
\noindent\textbf{Mirror Mechanism and Behavioural Cloning}
Humans possess a neural mechanism that allows them to map observed actions onto their own motor systems~\cite{bandura1977sociallearningtheory}.
This mirror mechanism extracts the underlying intention of action sequences and, afterwards, adapts and replicate the learned behaviour in similar situations.
For example, when watching someone stack blocks, humans learn goal-directed primitives, such as aligning, positioning, grasping, rather than just the actual hand and finger trajectories.

In computer science, a well-known attempt to capture the mirror mechanism is called {\em ``Behavioural Cloning''}~\cite{pomerleau1988alvinn}, where a policy $\policy$ learns parameters $\theta$ such that $\agent : \StateMDP \mapsto \Action$ by optimising: $\argmin_\theta \sum_{(s, a) \in \datasetTrain} \error(a, \agent(s))$, where $\datasetTrain$ is a set of teacher $\teacher$ demonstrations.
However, this simplification into supervised learning without capturing the compositional structure leads to a fundamental limitation: mimicry, which we explain next.

\vskip 3pt
\noindent\textbf{Forms of Imitation}
There are two forms of imitation~\cite{deshais2022generalized}: \textit{mimicry} and \textit{adaptive behaviour}.
Mimicry is associated with \textit{emulation}, i.e, agents retrieve the state in the demonstration that is most similar to the current state and then replay the associated action.
This causes \textit{behaviour-seeking mode}~\cite{swamy2021gap}, associated with two types of failures: compounding errors~\cite[Ch. 26]{russel2022artificial}, where small deviations cascade into large failures, and memorisation~\cite[Ch. 19]{russel2022artificial}, where agents retrieve stored actions rather than computing appropriate responses.
Adaptive behaviour generates novel responses while maintaining behavioural intent.
Agents understand not just what to do but also why, decomposing demonstrations into reusable primitives and compositional rules, enabling near-optimal responses to unseen states through recombination rather than retrieval.
Both forms of imitation are essential.
Mimicry works in controlled environments, while adaptive behaviour enables generalisation for more dynamic tasks where variation is the rule rather than the exception.

\vskip 3pt
\noindent\textbf{Limitation of Current Evaluation} \label{sec:sub:evaluation_limitation}
Current evaluation paradigms fail to measure true generalisation.
Current metrics, such as Average Episodic Reward (AER), hide critical failures through averaging.
An agent might achieve high AER by perfectly memorising most trajectories while failing on novel situations.
Furthermore, agentic applications (sequential decision problems, episodic in nature) create complex dependencies.
An agent starting from a state not present in the training dataset might reach familiar territory, then rely on memorisation. 
These metrics cannot distinguish this trajectory-level memorisation from compositional understanding.
Most critically, states not seen during training and testing create hidden failure points invisible during evaluation but catastrophic in deployment.
These limitations demand new paradigms that directly measure generalisation under a compositional lens.

\vskip 3pt
\noindent\textbf{Compositional Generalisation} \label{sec:sub:compositional}
Compositional generalisation, borrowed from linguistics~\cite{hupkes2020compositional}, provides the theoretical foundation for understanding adaptive capabilities.
Like humans understanding novel sentences through grammatical rules, agents must understand behavioural combinations through compositional principles.
Compositional generalisation refers to five forms through which a model can generalise:
\begin{enumerate*}[label=(\roman*)]
    \item \textit{systematicity}, which is generalisation via systematically recombining known parts and rules;
    \item \textit{productivity}, the ability to extend predictions beyond the length seen in training data (akin to machine learning generalisation);
    \item \textit{substitutivity}, which is generalisation via the ability to replace components with synonyms;
    \item \textit{localism}, whether the model compositions operate locally or globally; and
    \item \textit{overgeneralisation}, referring to the model's ability to pay attention to or being robust to exceptions.
\end{enumerate*}

In this work, we are particularly interested in the first three forms: systematicity, productivity, and substitutivity.
To exemplify them, we use an excerpt from~\citeauthor{kirk2023zeroshot}~\cite{kirk2023zeroshot}:
``[For a block-stacking environment], an example of \textit{systematicity} is the ability to stack blocks in new configurations once the basics of block-stacking are mastered.
Similarly, \textit{productivity} can be measured by how many blocks the agent can generalise to, the complexity of blocks, and the complexity of the stacking configuration.
\textit{Substitutivity} can be evaluated by the agent's ability to generalise to blocks of new colours, understanding that the new colour does not affect the physics of the block.''
Thus, systematicity, productivity, and substitutivity allow us to capture dimensions of generalisation that go beyond mere accuracy on held-out states.

\noindent\textbf{Related Approaches}
Beyond compositional generalisation, other noteworthy related approaches inspire \abbrev{}.
For example, \textbf{Goal-conditioned IL} policies $(\agent\mid s,g)$ enable flexible behaviour by conditioning on desired outcomes~\cite{kaelbling1993learning}.
Hindsight experience replay relabels failed trajectories with achieved goals~\cite{andrychowicz2017hindsight}.
However, goal-conditioning does not address how to deal with novel goal-context combinations.
Similarly, \textbf{Hierarchical and Skill-based IL} learns reusable behavioural components~\cite{niekum2015learning,sutton1999between}.
These approaches address primitive extraction but typically assume fixed composition rules.
\abbrev{} extends this by explicitly measuring compositional recombination across contexts.

\section{\problem} \label{sec:problem}

The path from mimicry to adaptability requires mathematical rigour that current IL frameworks lack.
This section lays \abbrev{}'s foundation by building on three components that transform compositional generalisation from intuition into a measurable quantity:
\textit{goals} enable trajectory-level success measurement, \textit{controllable contexts} provide systematic variation for testing, and \textit{deterministic transitions} isolate algorithmic performance from environmental randomness.

\subsection{Goal-conditioned Contextual MDPs} \label{sec:sub:cmdp}

The heart of \abbrev{} lies in a deceptively simple insight: standard MDPs cannot capture compositional structure because they conflate task definition with environmental dynamics.
When everything changes simultaneously, it is impossible to isolate what varies compositionally from what varies arbitrarily.
We propose Goal-conditioned Contextual MDPs (GCMDPs), extending contextual MDPs~\cite{hallak2015contextual} with explicit goal structure and context metrics.
\begin{definition}
    A \textit{Goal-conditioned Contextual MDP} is a tuple $\mdp_{gc} = \langle \StateMDP, \Action, \Context, \Goal,\allowbreak \Observation, \Transition, \projection, d \rangle$, where $\StateMDP$ is the state space, $\Action$ is the action space, $\Context$ is the set of controllable contexts, $\Observation$ is the set of observations, $\Goal$ is the set of goals, $\Transition : \Observation \times \Action \mapsto \Observation$ is a deterministic transition function, $\projection: \StateMDP \times\ \Context \mapsto \Observation$ is a projection function that maps state-context pairs to observations, and $d: \Context \times \Context \rightarrow \mathbb{R} \geq 0$ is a contextual distance metric.
\end{definition}
This formulation departs from standard problems in three critical ways.
First, by separating goals from rewards, we enable trajectory-level evaluation capturing compositional generalisation (as defined in Section~\ref{sec:background}) beyond local action accuracy.
Second, by using controllable contexts, GCMDPs avoid random variation, which obscures compositional structure.
Third, by introducing deterministic transition functions, it avoids stochasticity attribution problems.

Explicit goal specifications (declarative success criteria rather than numeric reward signals) provide the missing abstraction layer, defining success at the trajectory level to capture whether agents understand task structure beyond surface mimicry. 
\begin{definition}
    Let $\trajectory = \left( o_0, \ldots, o_n \right)$ be a trajectory, i.e., an ordered sequence of observations produced by the agent's interactions. $\Goal$ specifies success criteria for the agent and can be of one of the following types.
    \begin{enumerate*}[label=(\roman*)]
        \item \textit{terminal goals}: the agent succeeds if the final observation in a trajectory satisfies one of the designated goal states;
        \item \textit{landmark goals}: the agent succeeds when observations in a trajectory satisfy all states in one of the designated sets of goal states; or
        \item \textit{ordered landmark goals}: the agent succeeds when observations in a trajectory satisfy all states in one of the designated {\em sequences} of goal states (order preserving).
    \end{enumerate*}
\end{definition}
Goals enable deliberate failure analysis~\cite{amado2022goal}.
When agents fail, examining reached subgoals reveals whether failures stem from the inability to execute primitives~\cite{AmadoPereiraMeneguzzi2023}, to combine them properly, or to adapt combinations to novel contexts.

Controllable contexts expose generative parameters, creating variation.
For example, when agents fail on procedurally generated levels, it is hard to determine if they lack compositional understanding or encountered novel mechanics. Before we introduce them, let us say that two contexts $c$ and $c'$ are equivalent, in symbols $c \simeq c'$, if for all states $s$, we have that $\phi(s,c)=\phi(s,c')$.

\begin{definition}
    A context set $\Context$ is controllable when satisfying:
    \begin{enumerate*}[label=(\roman*)]
        \item \textit{distribution separation}, when for no $c \in \Context_{train}$ and $c' \in \Context_{test}$ do we have $c \simeq c'$;
        \item \textit{observation independence}, the agent should not assume that the projection function $\projection$ is injective (i.e., it does not assume equal observations in different contexts to be from the same state); and
        \item \textit{compositional structure}, where for all ${c \in \ContextTrain}$ and for all ${c' \in \ContextTest}$ such that $c\not\simeq c'$,  $0 < d(c, c') \leq \boundary_\Context$, where $\boundary_\Context = \max \{d(c,c') \mid  c \in  \ContextTrain, \, c' \in \ContextTest\}$.
    \end{enumerate*}
\end{definition}
In a block-stacking task, rather than random variation, controllable contexts allow control over problem characteristics, such as block count and colour.
For navigation, map layout, room connectivity, and obstacle placement become explicit parameters.
This enables systematic testing, varying single factors or specific combinations to reveal which compositional capabilities agents possess.
For symbolic contexts representable as vectors of properties, we advocate the \textit{Levenshtein} distance~\cite{lcvenshtcin1966binary}, which counts the number of insertion and deletion operations needed to transform between configurations. 
For example, transforming the specification of blocks from \textit{red, blue, green} to \textit{red, yellow, green} requires one deletion (\textit{blue}) and one insertion (\textit{yellow}), more accurately reflecting compositional changes, which Euclidean metrics struggle with.

Finally, deterministic transitions eliminate confounding stochasticity.
When agents fail in stochastic environments, we cannot distinguish between compositional limitations and unlucky events.
Consider navigation under unsafe conditions: failure might reflect inability to adapt (generalisation failure) or unstable random transitions that would fail regardless (environmental stochasticity).
Deterministic transitions eliminate this confounding factor.
The execution of an action in a state produces a unique outcome, ensuring that failures reflect the agent's limitations rather than chance.
This enables reproducible debugging, allowing researchers to trace exact failure points, identify violated compositional rules, and design targeted improvements.
Determinism transforms generalisation from a statistical property to a measurable one.

\subsection{Comparison with POMDPs}
A comparison between GCMDPs and Partially Observable MDPs (POMDPs) is warranted due to structural similarities.
POMDPs excel when agents must maintain beliefs about hidden states and perform Bayesian inference.
Yet, IL from expert demonstrations presents a different challenge: agents receive complete demonstrations with full state and context information.
The problem is not inferring hidden states but understanding compositional structure.
The projection $\projection : \StateMDP \times \Context \rightarrow \Observation$ captures how context shapes observation without hiding information.
This transparency enables systematic generalisation testing that is impossible under POMDPs' probabilistic framework.
When a GCMDP agent fails, we know it lacks compositional understanding, not state information, a far more tractable debugging problem.

\subsection{Measuring Compositional Generalisation}

Current IL metrics give us success rate but reveal nothing about the nature of that success.
An agent with $90\%$ accuracy may be simply replicating trajectories in familiar contexts but failing catastrophically on novel ones, or it may indeed possess robust compositional understanding, albeit with occasional execution errors.

Consider a block-stacking task where an agent learns to execute \texttt{stack} commands that build towers by placing blocks in sequence (e.g., \texttt{stack(red, blue)} places the red block on top of the blue one).
Suppose training includes blocks ``red'', ``blue'', and ``green'' in two-block configurations.
At test time, we might present novel challenges: a new block (\texttt{purple}), longer sequences (\texttt{red, blue, green, red}) or symmetrical configurations (\texttt{stack(blue, red)}).
Two agents with identical test accuracy might differ dramatically---one handles novel colours but fails on longer sequences, while the other shows the opposite pattern.
Traditional metrics cannot capture these distinctions.
Thus, we propose the \textit{generalisation boundary}, adapted from behavioural science~\cite{bandura1977sociallearningtheory}, that cuts through this ambiguity by explicitly measuring how far agents can deviate from training contexts while maintaining performance.
\begin{definition}
    Let $\Context$ be a set of controllable contexts, $p$ a minimum performance threshold, and $SR_\agent(d_c)$ the success rate at contextual distance $d_c$ (averaged over all contexts at that distance).
    We define the generalisation boundary as:
    $\boundary_\agent(p) = \sup\{d_c \in [0, \boundary_\Context] \mid \sr_\agent(d_c) \geq p\},$
    where $\boundary_\Context$ is the maximum contextual distance.
\end{definition}
This reveals generalisation profiles invisible to traditional evaluation.
Two agents with identical test accuracy might have vastly different boundaries: one maintaining perfect performance until sudden failure, while the other degrades gracefully as contextual distance increases.
High thresholds ($p = 0.9$) test robust systematicity, while lower thresholds ($p = 0.5$) reveal absolute limits of compositional understanding.
Combined with compositional generalisation, we can measure contextual distances to different test cases.
In our block-stacking example, we could change a block's colour to test substitutivity, add a block to test productivity, or reorder the blocks to test systematicity.
In addition, we can combine these forms of compositional generalisation to further test the agent.
Agents may succeed at different distances for different compositional dimensions.
An agent might handle substitutivity but fail systematicity, revealing specific compositional limitations.

\subsection{Suitable Environments and Benchmarks}
Not all environments enable compositional generalisation measurement.
In Procgen~\cite{cobbe2019procgen}, when agents fail on unseen levels, we cannot determine whether they lack compositional understanding or encountered entirely novel mechanics.
Compatible benchmarks require: 
\begin{enumerate*}[label=(\roman*)]
    \item fixed primitive semantics: a ``grasp'' action should function identically for red or blue blocks; environments changing primitives meanings conflate adaptation with relearning;
    \item isolated compositional dimensions: systematicity varies only structural combinations, productivity extends sequence length without novel primitives, substitutivity changes components with functional equivalents;
    \item interpretable failures: distinguishing nature of failures, perceptual (inability to recognise blocks), compositional (inability to plan sequences), or operational (inability to perform primitives); and 
    \item modular evaluation: primitives, compositional rules, and integrated behaviours separately.
\end{enumerate*}
Therefore, \abbrev{} calls for careful and well-thought-out benchmarks.
As \citeauthor{melanie2023howsmartaiare}~\cite{melanie2023howsmartaiare} argues, current AI evaluation often cannot distinguish memorisation from genuine understanding; compositional generalisation measurement inherits this challenge but, through careful benchmark design, offers an opportunity for deeper insight into agent capabilities.

\section{Future Research Landscape}

\abbrev{} opens research paths that extend beyond incremental improvements to IL.
These directions challenge fundamental assumptions about learning, adaptation, and intelligence while remaining grounded in achievable scientific progress.

\noindent\textbf{Compositional Generalization at Scale}:
Current benchmarks fail to measure compositional capabilities, conflating multiple dimensions or lacking structure entirely.
\abbrev{} demands specialised evaluation to reveal exactly which capabilities fail: 
systematicity, varying only in structural combinations; 
productivity, extending sequences without novel primitives; 
substitutivity, replacing objects with functional equivalents. 
Open-world deployment presents the ultimate challenge: agents must decompose novel situations, repurpose primitives, and recognise compositional limits. 
Research into compositional curricula that mimic human developmental progression from simple to complex behaviours could dramatically improve generalisation while revealing fundamental learning principles.

\noindent\textbf{Hybrid Architectures}:
\abbrev{} enables natural integration with complementary paradigms.
Foundation models provide semantic priors for primitive selection; planners excel at long-horizon reasoning.
Combining these yields agents that learn primitives via imitation but compose them using principled algorithms, addressing IL's inability to reason about novel combinations.
The challenge: preserving each paradigm's strengths while ensuring semantic alignment between learned primitives and symbolic representations.

\noindent\textbf{Safety and Ethics in Adaptive Systems}:
Creative recombination risks discovering harmful behaviours never demonstrated.
Compositional safety specifications must constrain entire combination spaces, not individual trajectories.
Behavioural contracts could guarantee safe primitive combinations—proving no navigation-grasping combination causes collision requires compositional algebra, not trajectory enumeration.
Ethical questions arise: should agent populations preserve behavioural diversity or converge? How do we ensure cross-cultural adaptation respects local norms?

\noindent\textbf{Multi-agent Compositional IL}:
Teams could share and recombine behavioural primitives, enabling collective repertoire expansion through social learning.
This mirrors cultural transmission mechanisms in human societies and opens questions about convergence versus diversity in behavioural populations.
Social learning theory shows how discoveries propagate through populations, suggesting mechanisms for efficient primitive sharing.

\noindent\textbf{Cognitive Science Synergies}:
Developmental psychology reveals a structured progression from mimicry to composition.
\citeauthor{kolb1983learning}'s experiential learning theory~\cite{kolb1983learning} suggests beneficial cycles: experiencing demonstrations, analysing patterns, extracting structures, and testing combinations, implying that agents should reflect rather than pure matching.
Behavioural science research on exploration-exploitation trade-offs can inform how agents balance exploring new combinations of primitives versus exploiting known compositions.
Social learning theory shows how discoveries propagate through populations, mirroring human cultural transmission.

The compositional structure from GCMDPs opens pathways we have not described here, such as:
multi-agent solutions using IL, where teams can share and recombine behavioural primitives, and implications for explainable AI through interpretable primitive decomposition.
By reconceptualising IL through the lens of compositionality, \abbrev{} transforms IL from a narrow technique for trajectory reproduction into a foundational paradigm for understanding how agent systems acquire, adapt, and transmit complex behaviours.
The breadth of these possibilities underscores that \abbrev{} is not merely an incremental advance but a research agenda that will reshape how we approach learning from demonstration.

\section{Conclusion}

This paper calls for a research agenda re-orientation in IL: from narrow mimicry toward lifelong adaptability through \problem{} (\abbrev).
By formalising Goal-conditioned Contextual MDPs and introducing compositional generalisation dimensions, we outline a foundation for agents that learn once and adapt everywhere.
\abbrev{} envisions architectures that allow for integrating imitation with planning, symbolic reasoning, and foundation models, supported by benchmarks that stress compositional generalisation (systematicity, productivity, and substitutivity).

The significance of this shift lies in enabling agents to operate beyond static demonstrations, adapt to unseen contexts, and recombine learned behaviours for novel goals.
These capabilities are critical for advancing autonomy in robotics, human-AI collaboration, and open-world systems where variability and uncertainty are the norm.
By embracing \abbrev{}, the agent community can establish IL as a cornerstone for next-generation AI, bridging data-driven learning with structured reasoning.
This will enable the realisation of more sophisticated cognitive learning techniques, paving the way for more robust, adaptable, and proactive agents.



\begin{acks}
This work was supported by UK Research and Innovation [grant number EP/S023356/1], in the UKRI Centre for Doctoral Training in Safe and Trusted Artificial Intelligence  (\url{www.safeandtrustedai.org}).
\end{acks}




\bibliographystyle{ACM-Reference-Format} 
\bibliography{sample}


\end{document}